\documentclass{article}
\usepackage{arxiv}

\usepackage[utf8]{inputenc} % allow utf-8 input
\usepackage[T1]{fontenc}    % use 8-bit T1 fonts
\usepackage{hyperref}
\usepackage{url}            % simple URL typesetting
\usepackage{booktabs}       % professional-quality tables
\usepackage{amsfonts}       % blackboard math symbols
\usepackage{nicefrac}       % compact symbols for 1/2, etc.
\usepackage{microtype}      % microtypography
\usepackage{mathtools}      % later include this.

\usepackage{algorithmic}
\usepackage{algorithm}
\usepackage{setspace}
\usepackage{color}
\usepackage{subfigure} 
\usepackage{amsmath}
\usepackage{graphicx}
\usepackage{multirow}
\usepackage{pifont}
\usepackage{marvosym}
\usepackage{array}
\usepackage{bm}
\usepackage{caption}
\usepackage{bbm}
\usepackage{natbib}

\title{Bi-tuning of Pre-trained Representations}

\author{%
    Jincheng Zhong\thanks{Equal contribution}, Ximei Wang\footnotemark[1], Zhi Kou, Jianmin Wang, Mingsheng Long (\Letter)\\
    School of Software, BNRist, Research Center for Big Data, Tsinghua University, China \\
		\{zjc19, wxm17, kz19\}@mails.tsinghua.edu.cn, \{mingsheng, jimwang\}@tsinghua.edu.cn \\
}

\begin{document}
	
	\maketitle
	\begin{abstract}
  %It is common within the deep learning community to first \textit{pre-train} a deep neural network from a large-scale dataset and then \textit{fine-tune} the pre-trained model to a specific downstream task. Supervised pre-training is the \emph{de facto} pre-training approach, which learns discriminative representations by exploiting the knowledge of labeled information. Currently, unsupervised pre-training has emerged as a highly promising approach to learning expressive representations by leveraging the knowledge of intrinsic structure underlying unlabeled data.
  
  It is common within the deep learning community to first \textit{pre-train} a deep neural network from a large-scale dataset and then \textit{fine-tune} the pre-trained model to a specific downstream task. Recently, both supervised and unsupervised pre-training approaches to learning representations have achieved remarkable advances, which exploit the discriminative knowledge of labels and the intrinsic structure of data, respectively. It follows natural intuition that both discriminative knowledge and intrinsic structure of the downstream task can be useful for fine-tuning, however, existing fine-tuning methods mainly leverage the former and discard the latter. A question arises: How to fully explore the intrinsic structure of data for boosting fine-tuning? In this paper, we propose Bi-tuning, a general learning framework to fine-tuning both supervised and unsupervised pre-trained representations to downstream tasks. Bi-tuning generalizes the vanilla fine-tuning by integrating two heads upon the backbone of pre-trained representations: a classifier head with an improved contrastive cross-entropy loss to better leverage the label information in an instance-contrast way, and a projector head with a newly-designed categorical contrastive learning loss to fully exploit the intrinsic structure of data in a category-consistent way. Comprehensive experiments confirm that Bi-tuning achieves state-of-the-art results for fine-tuning tasks of both supervised and unsupervised pre-trained models by large margins (\emph{e.g.}~10.7\% absolute rise in accuracy on \emph{CUB} in low-data regime).

\end{abstract}

	\iffalse

\bibliography{../ref/JZhong.bib}

\fi

\section{Introduction} 

% Modern efficient deep network training strategy gradually evolves into a two-stage style: pre-traine representations on a large-scale dataset like ImageNet~\cite{deng2009imagenet}, then fine-tune the model in specific downstream tasks~\cite{he2019momentum}. 
In the last decade, remarkable advances in deep learning have been witnessed in diverse applications across many fields, such as computer vision, robotic control, and natural language processing in the presence of large-scale labeled datasets. However, in many practical scenarios, we may have only access to a small labeled dataset, making it impossible to train deep neural networks from scratch. 
Therefore, it has become increasingly common within the deep learning community to first \textit{pre-train} a deep neural network from a large-scale dataset and then \textit{fine-tune} the pre-trained model to a specific downstream task. Fine-tuning requires fewer labeled data, enables faster training, and usually achieves better performance than training from scratch~\citep{He2019RethinkPretraining}.
This two-stage style of pre-training and fine-tuning lays as the foundation of various transfer learning applications. 
%In computer vision (CV), models pre-trained on ImageNet are widely employed to a variety of CV tasks such as image recognition~\cite{donahue_decaf:_2014}, object detection~\cite{girshick_rich_2014}, and semantic segmentation~\cite{he_mask_2017}, where impressive improvements are seen in many practical applications. In natural language processing (NLP), numerous NLP tasks, such as text classification~\cite{Sun2019fine-tuneBERT} and named entity recognition, gradually evolve into transferring from models pre-trained with a large-scale corpus, including BERT~\cite{devlin2018bert} and XLNet~\cite{Yang2019XLNet}.

%% 将pre-train和fine-tune合二为一
In the \emph{pre-training} stage, there are mainly two approaches to pre-train a deep model: supervised pre-training and unsupervised pre-training. 
Recent years have witnessed the success of numerous supervised pre-trained models, \emph{e.g.}~ResNet~\citep{he2016deep}, by exploiting the discriminative knowledge of labels on a large-scale dataset like ImageNet~\citep{deng2009imagenet}. 
% GoogleNet~\cite{Szegedy2015GoogleNet}, ResNet~\cite{he2016deep}, DenseNet~\cite{Huang2017DenseNet}, to name a few. 
%Given category information $y$ corresponding with each sample $\mathbf{x}$, supervised learning models are mostly pre-trained by applying a \emph{cross-entropy loss} on a labeled dataset $D$, which characterizes the \emph{conditional} distribution $p(y|\mathbf{x},D)$ of training data in general.
Meanwhile, unsupervised representation learning is recently changing the field of NLP by models pre-trained with a large-scale corpus, \emph{e.g.}~BERT~\citep{devlin2018bert} and GPT~\cite{radford2018improving}. %, BERT~\cite{devlin2018bert} and  XLNet~\cite{Yang2019XLNet}.
In computer vision, remarkable advances in unsupervised representation learning~\citep{wu2018unsupervised, he2019momentum, chen_simple_2020}, which exploit the intrinsic structure of data by contrastive learning \cite{Hadsell2006Dimensionality}, are also to change the field dominated chronically by supervised pre-trained representations. 
%Unsupervised learning models with a variety of pretext tasks (\emph{e.g.} instance discriminations, rotation) are generally pre-trained by applying a \emph{contrastive learning loss} \cite{gutmann_noise-contrastive_2010} on an unlabeled dataset. 
%Without supervision from the labels, this kind of generative models mainly model data representations based on the \emph{marginal} distribution $p(\mathbf{x}|D)$ of training data in essence. 
%Hence, these methods encourage unsupervised visual representations to concentrate more on local details in the training data, as shown in Figure~\ref{fig:motivation}. 

In the \emph{fine-tuning} stage, transferring a model from supervised pre-trained models has been empirically studied in~\cite{kornblith2019better}.
During the past years, several sophisticated fine-tuning methods were proposed, including L2-SP~\citep{Li2018L2SP}, DELTA~\citep{Li2019Delta} and BSS~\citep{Chen2019BSS}.
%During the past years, a few fine-tuning methods have been proposed to exploit the inductive bias of pre-trained models: L2-SP~\cite{Li2018L2SP} drives the weight parameters of target task to the pre-trained values by imposing L2 constraint; DELTA~\cite{Li2019Delta} computes channel-wise discriminative knowledge to reweight the feature map regularization with an attention mechanism; BSS~\cite{Chen2019BSS} penalizes smaller singular values to suppress untransferable spectral components. 
%These strategies work for supervised pre-trained models since both of the pre-trained models and the models after fine-tuning focus on capturing global category-structure, though them may still fall short when labeled dataset is limited.
These methods focus on leveraging the discriminative knowledge of labels by a {cross-entropy loss} and the implicit bias of pre-trained models by a regularization term.
However, the intrinsic structure of data in downstream task is generally discarded during fine-tuning.
%% 快速破题
%% why we need deep tuning? 不足和难点
%%
Further, we empirically observed that unsupervised pre-trained representations focus more on the intrinsic structure, while supervised pre-trained representations explain better on the label information (Figure~\ref{fig:motivation}). This possibly implies that fine-tuning unsupervised pre-training representations is may be more difficult \cite{he2019momentum}.

Regarding to the success of supervised and unsupervised pre-training approaches, it follows a natural intuition that both \emph{discriminative knowledge} and \emph{intrinsic structure} of the downstream task can be useful for fine-tuning. A question arises: How to fully explore the intrinsic structure of data for boosting fine-tuning? To tackle this major challenge of deep learning, we propose \textbf{Bi-tuning}, a general learning framework to fine-tuning both supervised and unsupervised pre-trained representations to downstream tasks. Bi-tuning generalizes the vanilla fine-tuning by integrating two heads upon the backbone of pre-trained representations: 
\begin{itemize}
	\item A classifier head with an improved contrastive cross-entropy loss to better leverage the label information in an instance-contrast way, which is the dual view of the vanilla cross-entropy loss and is expected to achieve a more compact intra-class structure.
	\item A projector head with a newly-designed categorical contrastive learning loss to fully exploit the intrinsic structure of data in a category-consistent way, resulting in a more harmonious cooperation between the supervised and unsupervised fine-tuning mechanisms.
\end{itemize}
As a general fine-tuning framework, Bi-tuning can be applied with a variety of backbones without any additional assumptions.
Comprehensive experiments confirm that Bi-tuning achieves state-of-the-art results for fine-tuning tasks of both supervised and unsupervised pre-trained models by large margins (\emph{e.g.}~10.7\% absolute rise in accuracy on \emph{CUB} in low-data regime). We justify through ablation studies the effectiveness of the proposed two-heads fine-tuning architecture with their novel loss functions.

\section{Related Work} 
\subsection{Pre-training}
%For the first stage, there are mainly two approaches to \textit{pre-train} a model: supervised pre-training and unsupervised pre-training. 
 During the past years, supervised pre-trained models achieve impressive advances by exploiting the inductive bias of label information on a large-scale dataset like ImageNet~\citep{deng2009imagenet}, such as GoogleNet~\citep{Szegedy2015GoogleNet}, ResNet~\citep{he2016deep}, DenseNet~\citep{Huang2017DenseNet}, to name a few. 
%Given category information $y$ corresponding with each sample $\mathbf{x}$, supervised learning models are mostly pre-trained by applying a \emph{cross-entropy loss} on a labeled dataset $D$, which characterizes the \emph{conditional} distribution $p(y|\mathbf{x},D)$ of training data in general.
Meanwhile, unsupervised representation learning is recently shining in the field of NLP by models pre-trained with a large-scale corpus, including GPT~\citep{radford2018improving}, BERT~\citep{devlin2018bert} and  XLNet~\citep{Yang2019XLNet}.
Even in computer vision, recent impressive advances in unsupervised representation learning~\citep{wu2018unsupervised, he2019momentum, chen_simple_2020}, which exploit the inductive bias of data structure, are shaking the long-term dominated status of representations learned in a supervised way. Further, a wide range of handcrafted pretext tasks have been proposed for unsupervised representation learning, such as relative patch prediction \citep{doersch2015unsupervised}, solving jigsaw puzzles \citep{noroozi2016unsupervised}, colorization \citep{zhang2016colorful}, etc.

\subsection{Contrastive Learning}
 Specifically, various unsupervised pretext tasks are based on some forms of contrastive learning, in which the instance discrimination approach \citep{wu2018unsupervised,he2019momentum,chen_simple_2020} is one of the most general forms. Other variants of contrastive learning methods include contrastive predictive learning (CPC) \citep{oord2018representation} and colorization contrasting \citep{tian2019contrastive}.
%Meanwhile, the instantiations of queries and keys in contrastive learning depend on the context of the pretext task. 
Recent advances of deep contrastive learning benefit from contrasting positive keys against \emph{very large} number of negative keys. Therefore, how to efficiently generate keys becomes a fundamental problem in contrastive learning. To achieve this goal, \cite{doersch2017multi} explored the effectiveness of in-batch samples, \cite{wu2018unsupervised} proposed to use a memory bank to store all representations of the dataset, \cite{he2019momentum} further replaced a memory bank with the momentum contrast (MoCo) to be memory-efficient, and \cite{chen_simple_2020} showed that a brute-force huge batch of keys works well.

\subsection{Fine-tuning}
Fine-tuning a model from supervised pre-trained models has been empirically explored in~\cite{kornblith2019better} by launching a systematic investigation with grid search of the hyper-parameters.
During the past years, a few fine-tuning methods have been proposed to exploit the inductive bias of pre-trained models: L2-SP~\citep{Li2018L2SP} drives the weight parameters of target task to the pre-trained values by imposing L2 constraint based on the inductive bias of parameter; DELTA~\citep{Li2019Delta} computes channel-wise discriminative knowledge to reweight the feature map regularization with an attention mechanism based on the inductive bias of behavior; BSS~\citep{Chen2019BSS} penalizes smaller singular values to suppress untransferable spectral components based on the inductive bias of singular value.
Other fine-tuning methods including learning with similarity preserving \citep{kang2019structure} or learning without forgetting \citep{li2017learning} also work well on some downstream classification tasks. However, the existing fine-tuning methods mainly focus on leveraging the knowledge of the target label with a {cross-entropy loss}. Intuitively, encouraging a model to capture the label information and intrinsic structure simultaneously may help the model transition between the upstream unsupervised models with the downstream classification tasks. In natural language processing, GPT~\citep{radford2018improving,radford2019language} has employed a strategy that jointly optimizes unsupervised training criteria while fine-tuning with supervision. However, we empirically found that trivially following this kind of force-combination between supervised learning loss and unsupervised contrastive learning loss is beneficial but limited. The plausible reason is that these two losses will contradict with each other and result in an extremely different but not discriminative feature structure compared to that of the supervised cross-entropy loss (See Figure~\ref{fig:motivation}).
% It remains unclear how to leverage the knowledge of unsupervised representations to downstream classification tasks, not to mention designing a general transfer learning approach to fine-tuning both supervised and unsupervised pre-trained models.

%When fine-tuning from an unsupervised pretrained model, the parameters change drastically, making it impossible to simply apply the fine-tuning methods that perform well in supervised pretraining setting. Therefore, it is \textit{a blank area} in the deep learning community and CORE is exactly such a missing technique we desire.

\section{Preliminaries}
It is noteworthy that the spirits of contrastive learning actually can date back very far \citep{becker1992self,Hadsell2006Dimensionality,gutmann_noise-contrastive_2010}. The key idea is to maximize the likelihood of the distribution $p(\mathbf{x}|D)$ contrasting to the artificial noise distribution $p_n(\mathbf{x})$, also known as noise-contrastive estimation (NCE). Later, \cite{goodfellow2014generative} pointed out the relations between generative adversarial networks and noise-contrastive estimation.
Meanwhile, \citep{oord2018representation} revealed that contrastive learning is related to mutual information between a query and the corresponding positive key, which is known as InfoNCE.  Considering a query $q$ with a large key pool $\{k_0,k_1,k_2,\cdots,k_{K}\}$ where $K$ is the number of keys, this kind of non-parametric form~\citep{oord2018representation,wu2018unsupervised} of {contrastive loss} can be defined as follows:
\begin{equation}\label{infonce}
L_{ \rm{InfoNCE}}= -\log\frac{\exp(q\cdot k_{+}/\tau)}{\sum_{i=0}^{K}\exp(q\cdot k_{i}/\tau)},
\end{equation}
where $\tau$ is the hyper-parameter for temperature scaling. Intuitively, contrastive learning can be defined as a query-key pair matching problem, in which a contrastive loss is a $K$-way cross-entropy loss to distinguish $k_+$ from a large key pool.  From this perspective, a contrastive loss is to maximize the similarity between the query and the corresponding positive key $k_+$. 

	\iffalse
\bibliography{../ref/JZhong.bib}
\fi

%1. Pretrained Representations

%2. Vanilla fine-tuning

%3. clssifier head  improve标准的finetune: 
%   CE + CCE
%   好处 logit space
%  横着算和竖着算 
% class-wise  and point-wise
%   dual format 
%   在loss层面的不足
% label information
% CE 和 CCE的区别和联系

% 4. project head:
% not enough 
% Hinton 不能做无监督fine-tune
% +projector inspired by unsupervised pre-trained
%  无监督的方法 ==> 有监督
%  Category Contrastive Loss CCL
%  navie combine CL 是不行的==> 矛盾
% intrinsic structure
% CL 和 CCL的区别和联系

% Bi-tuning 、Deep-Tuning、Dual-Tuning

%In the proposed Bi-tuning, we fine-tune an encoder $f(\cdot)$ that is from unsupervised pretrained model through two contrastive processes besides the cross-entropy loss $L_{\text{CE}}$. Projector $g(\cdot)$ embeds encoded query $h^q$ into a latent metric space and classifier $\textbf{w}$ embeds $h^q$ to a logit space. The keys $z^k$ and $h^k$ are from the latent key pool and hidden key pool respectively, which can be produced by key generating mechanism. (i) In the latent \emph{metric space}, we tend to maximize the similarity between query $z^q$ and positive keys (in same category) contrasted with the keys from different categories by minimizing the \emph{category-aware} contrastive loss $L_{\text{CCL}}$; (ii) In the \emph{logit space}, similar contrastive process is executed among query logit sBi-tunings ($\textbf{w}_{y^q} {\bf h}^q$ and $\textbf{w}_{y^q} {\bf h}^k$), by minimizing \emph{manifold-aware} cross-entropy loss $L_{\text{MCE}}$.

\section{Method}

\subsection{Pre-trained Representations} 
Bi-tuning is a general learning approach to fine-tuning both supervised and unsupervised representations. Without any additional assumptions, the pre-trained feature encoder $f(\cdot)$ can be various network backbones according to the downstream tasks, including ResNet~\citep{he2016deep} and DenseNet~\citep{Huang2017DenseNet} for supervised pre-trained models, and MoCo~\citep{he2019momentum} and SimCLR~\citep{chen_simple_2020} for unsupervised pre-trained models.
% and GPT~\citep{radford2018improving}, and BERT~\citep{devlin2018bert} for NLP tasks.
Given a query sample $\mathbf{x}^{\rm q}_i$, we can first utilize a pre-trained feature encoder $f(\cdot)$ to extract its pre-trained representation as $\mathbf{h}^{\rm q}_i=f(\mathbf{x}^{\rm q}_i)$.
% \begin{equation}\label{Pre-trained}
% \mathbf{h}^{\rm q}_i=f(\mathbf{x}^{\rm q}_i).
% \end{equation} 

\subsection{Vanilla Fine-tuning} 
Given a pre-trained representation $\mathbf{h}^{\rm q}_i$, a fundamental step of vanilla fine-tuning is to feedforward the representation $\mathbf{h}^{\rm q}_i$ into a $C$-way classifier $g(\cdot)$, in which $C$ is the number of categories for the downstream classification task. Denote the parameters of the classifier $g(\cdot)$ as $\textbf{W} = \left[\textbf{w}_1, \textbf{w}_2,\cdots, \textbf{w}_C\right]$, where $\textbf{w}_j$ corresponds to the parameter for the $j$-th class. Denote the training dataset of the downstream task as $\{(\mathbf{x}^{\rm q}_i, {y}^{\rm q}_i)\}_{i=1}^N$. $\textbf{W}$ can be updated by optimizing a standard cross-entropy (CE) loss as
\begin{equation}\label{CE}
	L_{\rm {CE}}= -\sum_{i=1}^N\log \frac{\exp(\mathbf{w}_{y_i}\cdot\textbf{h}_i^{\rm q})}{\sum_{j=1}^{C}\exp(\mathbf{w}_{j}\cdot\textbf{h}_i^{\rm q})}.
\end{equation}

\begin{figure*}[tbph]
	\centering
	\vspace{0pt}
	\includegraphics[width=\textwidth]{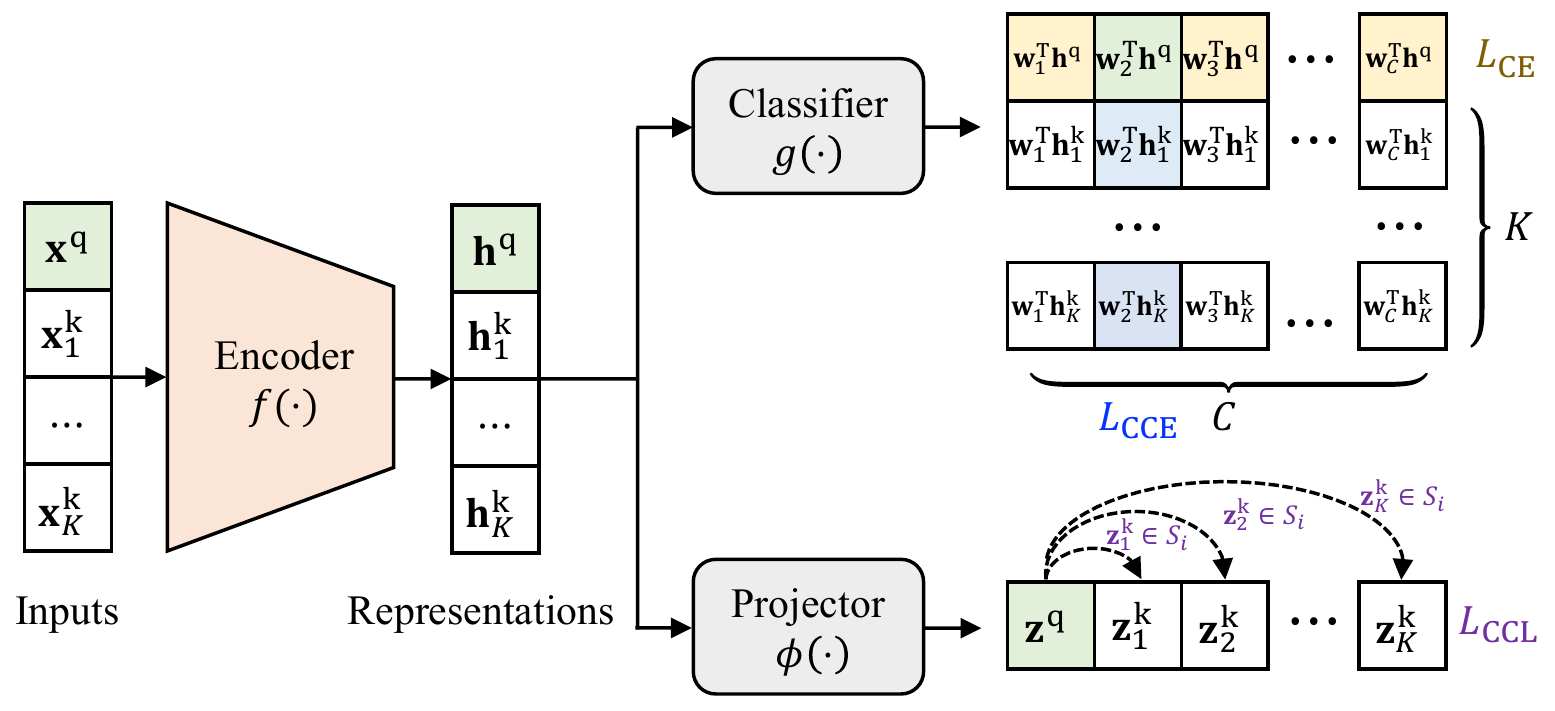}
	\vspace{-10pt}
	\caption{The architecture of the proposed \textbf{Bi-tuning} approach, which includes an encoder for pre-trained representations, a classifer head and a projector head. Bi-tuning enables a dual fine-tuning mechanism: a contrastive cross-entropy loss (CCE) on the classifier head to exploit label information and a categorical contrastive learning loss (CCL) on the projector head to model the intrinsic structure.}
	\label{fig:Structure}
	\vspace{-5pt}
\end{figure*}

\subsection{Contrastive Cross-Entropy Loss on Classifier Head}
\label{sec:fcl}
%3. clssifier head  improve标准的finetune: 
%   CE + CCE
%  dual format 
%  logit space
%  好处 
%  横着算和竖着算 
%  class-wise  and point-wise
%  在loss层面的不足
%  label information
%  CE 和 CCE的区别和联系
%Besides of adapting the InfoNCE into a category-aware contrastive learning loss, we further improve the cross-entropy into a \emph{manifold-aware cross-entropy} (MCE). 
%（点，类）pair
%类上竞争，点突出
%点上竞争，类突出
From another perspective, the cross-entropy loss of vanilla fine-tuning on a given dataset with $N$ instance-class pair $(\mathbf{x}^{\rm q}_i, {y}^{\rm q}_i)$ can be regarded as a class-wise championship, \emph{i.e.}, the prediction that is the same as the ground-truth class of each instance is expected to win the game. 
To further exploit the label information of the downstream task, we propose a novel contrastive cross-entropy loss $L_{\rm{CCE}}$ on the classifier head via the dual view of cross-entropy loss. 
Similarly, $L_{\rm{CCE}}$ can be seen as an instance-wise championship, \emph{i.e.}, the prediction belongs to the nearest instance towards the prototype of each class is expected to win the game.
Similar to CE loss, $L_{\rm{CCE}}$ can be formulated as
\begin{equation}\label{CCE}
L_{\rm{CCE}}= -\sum_{i=1}^N\sum_{k=0}^K\mathbbm{1}({\mathbf{h}_k^{\rm k} \in S_i})\log \frac{\exp(\mathbf{w}_{y_i}\cdot \mathbf{h}_i^{\rm q}/\tau)}{\sum_{j=0}^{K}\exp(\mathbf{w}_{y_i}\cdot \mathbf{h}_j^{\rm k}/\tau)},
\end{equation}
where $K$ is the size of candidate keys, $S_i$ is the positive key set for example $i$, and $\mathbbm{1}(\cdot) \in \{0,1\}$ is an indicator function that values $1$ if and only if the input condition holds. Note that, ${\bf h}^{\rm k}$'s are samples from the hidden key pool produced by the key generating mechanism. Though Bi-tuning is general for it, we adopt the key generating approach in Momentum Contrast (MoCo) \citep{he2019momentum} as our default one due to its simplicity, high-efficacy, and memory-efficient implementation.
As intuitively illustrated in Figure~\ref{fig:Structure}, column-wised $L_{\rm CCE}$ operates loss computation along the bank dimension whose size is $K+1$ while row-wised $L_{\rm CE}$ performs along the class dimension with a size of $C$. 
By encouraging instances in the training dataset to approach towards their corresponding class prototypes, $L_{\rm{CCE}}$ tends to achieve a more compact intra-class structure than the vanilla fine-tuning.

\subsection{Categorical Contrastive Learning Loss on Projector Head}
\label{sec:ccl}
%Consider an encoded query q and a set of encoded sam- ples {k0, k1, k2, ...} that are the keys of a dictionary. As- sume that there is a single key (denoted as k+) in the dic- tionary that q matches. A contrastive loss [29] is a function whose value is low when q is similar to its positive key k+ and dissimilar to all other keys (considered negative keys for q). With similarity measured by dot product, a form of a contrastive loss function, called InfoNCE [46], is consid- ered in this paper:
%Meanwhile, the assumption of unsupervised contrastive learning that there is only a single key in the dictionary that query matches and implicitly considers every sample as an individual class \cite{li2017learning,he2019momentum} is not reasonable in supervised fine-tuning.
% 4. project head:
% not enough 
% Hinton 不能做无监督fine-tune
% +projector inspired by unsupervised pre-trained
%  无监督的方法 ==> 有监督
%  Category Contrastive Loss CCL
%  navie combine CL 是不行的==> 矛盾
% intrinsic structure
% CL 和 CCL的区别和联系

Previously, we propose an improved-version of vanilla fine-tuning on the classifier head to fully exploit label information. However, this kind of loss function design may still fall short in capturing the intrinsic structure. 
Inspired by the remarkable success of unsupervised pre-training which also aims at modeling the intrinsic structure in data, we first introduce a projector $\phi(\cdot)$ which is usually off the shelf to embed a pre-trained representation $\mathbf{h}^{\rm q}_i$ into a latent metric space as $\mathbf{z}^{\rm q}_i$. 
However, the standard contrastive learning loss (InfoNCE) defined in Eq.~\eqref{infonce} assumes that there is a \textit{single} key $k_+$ in the dictionary to match the given query $q$, which implicitly requires every instance to belong to an individual class. 
If we simply apply InfoNCE loss on the labeled downstream dataset, it will result in an extremely different but not discriminative feature structure compared with that of the supervised cross-entropy loss, making the classifier struggle.
Obviously, this dilemma reveals that the naive combination of the supervised cross-entropy loss and the unsupervised contrastive loss is not an optimal solution for fine-tuning, which is also backed by our experiments in Table~\ref{table:result-unsup}. 

To capture the label information and intrinsic structure simultaneously, we propose a novel categorical contrastive loss $L_{\rm CCL}$ on the projector head based on the following hypothesis: when we fine-tune a pre-trained model to a downstream task, it is reasonable to regard other keys in the same class as the positive keys that the query matches. In this way, $L_{\rm CCL}$ expands the scope of positive keys to \textit{a set of instances} instead of a \textit{single one}, resulting in a more harmonious cooperation between the supervised and unsupervised learning mechanisms.
Similar to the format of InfoNCE loss, $L_{\rm CCL}$ is defined as 
\begin{equation}\label{CCL}
	L_{\rm CCL}= -\sum_{i=1}^N \sum_{k=0}^{K}\mathbbm{1}({\mathbf{z}_k^{\rm k} \in S_i}) \log\frac{\exp(\mathbf{z}_i^{\rm q}\cdot \mathbf{z}_k^{\rm k}/\tau)}{\sum_{j=0}^{K}\exp(\mathbf{z}_i^{\rm q}\cdot \mathbf{z}_j^{\rm k}/\tau)},
\end{equation}
where the notations are identical to Eq.~\eqref{CCE}, as well as the positive key set. Note that, the second sum is over all candidate keys, indicating that there may be more than one positive key for a single query, \emph{i.e.}, the inequality that $|S_i|\ge 1$ holds.
%$S_i$ can be straightforwardly defined as all keys related to the same class are considered to be the positive set for all training examples in that category. 
%Since the InfoNCE loss in Eq.~\eqref{infonce} is incapable of handling the case where the key samples are known to belong to a certain category, we propose a \emph{category-aware contrastive learning} (CCL) loss that intuitively extends InfoNCE loss to a variant that can handle the supervised case. 
%\paragraph{Positive key set of a query}In classic classification paradigm. Postive set of a query can be defined straghtforward, that all keys related to the same category are considered to be the positive set for all training examples in that category. The specific definition of postive sets depends on retrained task.
%When the dataset is small, it would be hard for the backbone to be  thoroughly adapted with the cross-entropy loss.
%So we first propose a loss function that lies in the form of contrastive loss but could train the model with supervised information.

\subsection{Bi-tuning Framework}
Finally, we reach a novel approach to fine-tuning both supervised and unsupervised representations, \emph{i.e.} the \textbf{Bi-tuning} framework, which jointly optimizes the standard cross-entropy loss, the contrastive cross-entropy loss for classifier head and the categorical contrastive learning loss for projector head in an end-to-end deep architecture. The overall loss function of Bi-tuning can be formulated as follows:
\begin{equation}\label{final}
\min_{\Theta} L_{\rm {CE}} + L_{\rm{CCE}} + L_{\rm {CCL}},
\end{equation}
where $\Theta$ denotes the set of all parameters of the backbone, the classifier head and the projector head.
Specifically, since the magnitude of the above loss terms is comparable, we empirically find that there is no need to introduce any extra hyper-parameters to trade-off them. This simplicity makes Bi-tuning easy to be applied to different datasets or tasks. The full portrait of Bi-tuning is shown in Figure~\ref{fig:Structure}.

\section{Experiments}
% datasets
%In this section, we prove our proposed \textit{Bi-tuning} is a unified, effective framework empirically. 
%In this section, we empirically validate \textit{Bi-tuning} on both supervised pre-trained and unsupervised pre-trained models, as well as detailed analysis on the size of key pool, the dimension of latent space, pre-trained approaches, feature visualization and collaborative effects.

%from two orthogonal perspectives: First, we evaluate four visual recognition tasks with various dataset scales. Second, we fine-tune from representations obtained from both supervised pre-training and unsupervised pre-training.

%\paragraph{Implement Details} 
We follow the common fine-tuning principle described in \cite{yosinski2014transferable}, replacing the last task-specific layer in the classifier head with a randomly initialized fully connected layer whose learning rate is $10$ times of that for pre-trained parameters. Meanwhile, the projector head is set to be another randomly initialized fully connected layer. For the key generating mechanisms, we follow the style in \cite{he2019momentum}, employing a momentum contrast branch with a default momentum coefficient $m=0.999$ and two cached queues both normalized by their L2-norm~\citep{wu2018unsupervised} with dimensions of $2048$ and $128$ respectively. For each task, the best learning rate is selected by cross-validation under a 100\% sampling rate and applied to all four sampling rates. Queue size $K$ is set as $8,16,24,32$ for each category according to the dataset scales respectively. Other hyper-parameters in Bi-tuning are fixed for all experiments.  The temperature $\tau$ in Eq. \eqref{CCE} and Eq. \eqref{CCL}  is set as $0.07$~\citep{wu2018unsupervised}. The trade-off coefficients between these three losses are kept as $1$  since the magnitude of the loss terms is comparable. All tasks are optimized using SGD with a momentum $0.9$.
%NLP tasks are optimized by Adam. 

%Both results for supervised pre-training and unsupervised pre-training are presented.

\begin{table*}[!t]
	%\vspace{-10pt}
	\addtolength{\tabcolsep}{2pt}
	\centering
	\small
	\vskip 0.1in
	% \resizebox{\textwidth}{!}{%
	\caption{Top-1 accuracy on various datasets using ResNet-50 by \textit{supervised pre-training}.}
	
	\begin{tabular}{llcccc}
		% {p{2.1cm}<{\centering}|c|p{1.5cm}<{\centering }p{1.5cm}<{\centering} p{1.5cm}<{\centering} p{1.5cm}<{\centering}}
		\toprule
		\multirow{2}*{Dataset} & \multirow{2}*{Method} & \multicolumn{4}{c}{Sampling Rates} \\
		& & $25\%$ & $50\%$ & $75\%$ & $100\%$ \\
		\midrule
		\multirow{5}*{CUB}      & Fine-tuning (baseline)        & $61.36\pm$0.11          & $73.61\pm$0.23          & $78.49\pm$0.18          & $80.74\pm$0.15          \\
		& L2SP~\citep{Li2018L2SP}      & $61.21\pm$0.19         & $72.99\pm$0.13          & $78.11\pm$0.17          & $80.92\pm$0.22              \\
		& DELTA~\citep{Li2019Delta}    & $62.89\pm$0.11         & $74.35\pm$0.28          & $79.18\pm$0.24          & $81.33\pm$0.24              \\
		& BSS~\citep{Chen2019BSS}      & $64.69\pm$0.31         & $74.96\pm$0.21          & $78.91\pm$0.15          & $81.52\pm$0.11          \\
		& \textbf{Bi-tuning}          & $\textbf{67.47}\pm$0.08 & $\textbf{77.17}\pm$0.13 & $\textbf{81.07}\pm$0.09 & $\textbf{82.93}\pm$0.23 \\
		\midrule
		\multirow{5}*{Cars}     & Fine-tuning  (baseline)         & $56.45\pm$0.21          & $75.24\pm$0.17          & $83.22\pm$0.17          & $86.22\pm$0.12          \\
		& L2SP~\citep{Li2018L2SP}      & $56.29\pm$0.21              & $75.62\pm$0.32              & $83.60\pm$0.13              & $85.85\pm$0.12              \\
		& DELTA~\citep{Li2019Delta}    & $58.74\pm$0.23              & $76.53\pm$0.08              & $84.53\pm$0.29              & $86.01\pm$0.37              \\
		& BSS~\citep{Chen2019BSS}      & $59.74\pm$0.14          & $76.78\pm$0.16          & $85.06\pm$0.13          & $87.64\pm$0.21          \\
		& \textbf{Bi-tuning}         & $\textbf{66.15}\pm$0.20 & $\textbf{81.10}\pm$0.07 & $\textbf{86.07}\pm$0.23 & $\textbf{88.47}\pm$0.11  \\
		
		\midrule
		\multirow{5}*{Aircraft}& Fine-tuning (baseline)           & $51.25\pm$0.18          & $67.12\pm$0.41          & $75.22\pm$0.09          & $79.18\pm$0.20              \\
		& L2SP~\citep{Li2018L2SP}       & $51.07\pm$0.45          & $67.46\pm$0.22          & $75.06\pm$0.45          & $79.07\pm$0.21                       \\
		& DELTA~\citep{Li2019Delta}     & $53.71\pm$0.30          & $68.51\pm$0.24          & $76.51\pm$0.55          & $80.34\pm$0.14                       \\
		& BSS~\citep{Chen2019BSS}       & $53.38\pm$0.22          & $69.19\pm$0.18          & $76.39\pm$0.22          & $80.83\pm$0.32          \\
		& \textbf{Bi-tuning}          & $\textbf{58.27}\pm$0.26 & $\textbf{72.40}\pm$0.22 & $\textbf{80.77}\pm$0.10 & $\textbf{84.01}\pm$0.33 \\
		\bottomrule
		
	\end{tabular}
	
	\label{table:result-sup}
	
\end{table*}

\begin{table*}[!t]
	%\vspace{-10pt}
	\small
	\addtolength{\tabcolsep}{2pt}
	\centering
	\vskip 0.1in
	% \resizebox{\textwidth}{!}{%
	\caption{Top-1 accuracy on \textbf{COCO-70} dataset using DenseNet-121 by \textit{supervised pre-training}.}
	
	\begin{tabular}{llcccc}
		% {p{2.1cm}<{\centering}|c|p{1.5cm}<{\centering }p{1.5cm}<{\centering} p{1.5cm}<{\centering} p{1.5cm}<{\centering}}
		\toprule
		\multirow{2}*{Dataset} & \multirow{2}*{Method} & \multicolumn{4}{c}{Sampling Rates} \\
		& & $25\%$ & $50\%$ & $75\%$ & $100\%$ \\
		\midrule
		
		\multirow{5}*{COCO-70}     
		& Fine-tuning (baseline) & $80.01\pm$0.25 &$82.50\pm$0.25 & $83.43\pm$0.18 & $84.41\pm$0.22  \\
		& L2SP~\citep{Li2018L2SP}        & $80.57\pm$0.47 &$80.67\pm$0.29 & $83.71\pm$0.24 & $84.78\pm$0.16  \\
		& DELTA~\citep{Li2019Delta}       & $76.39\pm$0.37\ &$79.72\pm$0.24 & $83.01\pm$0.11 & $84.66\pm$0.08  \\
		& BSS~\citep{Chen2019BSS}         & $77.29\pm$0.15 &$80.74\pm$0.22 & $83.89\pm$0.09 & $84.71\pm$0.13 \\
		& \textbf{Bi-tuning} & $\textbf{80.68}\pm$0.23 & $\textbf{83.48}\pm$0.13 & $\textbf{84.16}\pm$0.05 & $\textbf{85.41}\pm$0.23  \\
		\bottomrule
		
	\end{tabular}
	
	\label{table:result-coco}
	
\end{table*}

\subsection{Bi-tuning Supervised Pre-trained Representations}
\label{Exp-low-data-section}
%We first verify \textit{Bi-tuning} under low-data regime, which is very common in real applications. In this subsection 
\textbf{Standard benchmarks.} We first verify our framework on three fine-grained classification benchmarks: \textbf{CUB-200-2011}~\citep{cub200-WelinderEtal2010} (with $11788$ images for $200$ bird species), \textbf{Stanford Cars}~\citep{cars-KrauseStarkDengFei-Fei_3DRR2013} (containing $16185$ images of $196$ classes of cars) and \textbf{FGVC Aircraft} ~\citep{aircraft-maji2013fine} (containing $10000$ samples $100$ different aircraft variants). For each benchmark, we create four configurations which randomly sample $25\%, 50\%, 75\%$, and $100\%$ of training data for each class respectively, to reveal the detailed effect while fine-tuning to different data scales.
We choose recent fine-tuning technologies: {L2-SP}~\citep{Li2018L2SP}, {DELTA}~\citep{Li2019Delta}, and the state-of-the-art method {BSS}~\citep{Chen2019BSS},  as competitors of {Bi-tuning} while regarding vanilla fine-tuning as a baseline. Note that vanilla fine-tuning is a strong baseline when sufficient data is provided. Results are averaged over $5$ trials. As shown in \ref{table:result-sup}, Bi-tuning outperforms all competitors significantly across all three benchmarks by large margins (\emph{e.g.} 10.7\% absolute rise on \emph{CUB} with a sampling rate of 25\%). Note that even under $100\%$ sampling rate, Bi-tuning still outperforms others. 
%with standard ResNet-50~\citep{he2016deep} supervised pre-trained on ImageNet.

\textbf{Large-scale benchmarks.} Previous fine-tuning methods mainly focus on improving performance under low-data regime paradigms. We further extend Bi-tuning framework to large-scale paradigms. We use annotations of COCO dataset ~\citep{10.1007/978-3-319-10602-1_48} to construct a large-scale classification dataset, cropping object with padding for each image and removing minimal items (with height and width less than 50 pixels), resulting a large-scale dataset containing 70 classes with more than 1000 images per category. The scale is comparable to ImageNet in terms of the number of samples per class. On this constructed large-scale dataset named COCO-70, Bi-tuning is also evaluated under four sampling rate configurations. Since even 25\% sampling rates of COCO-70 is much larger than each benchmark in Section~\ref{Exp-low-data-section}, previous fine-tuning competitors show micro contributions to these paradigms. Results in Table~\ref{table:result-coco} reveal that Bi-tuning brings general gains for all tasks, even provided with sufficient training data. We hypothesize that the intrinsic structure introduced by Bi-tuning contributes substantially.
%Here we choose DenseNet-121 for supervised pre-training setting and ResNet-50 unsupervised pre-trained via MoCo as backbones. 

\begin{table*}[h]
	%\vspace{-10pt}
	\small
	\addtolength{\tabcolsep}{2pt}
	\centering
	\vskip 0.1in
	% \resizebox{\textwidth}{!}{%
	\caption{Top-1 accuracy on various datasets using ResNet-50 \textit{unsupervisedly pre-trained} by MoCo.}

	\begin{tabular}{llcccc}
		% {p{2.1cm}<{\centering}|c|p{1.5cm}<{\centering }p{1.5cm}<{\centering} p{1.5cm}<{\centering} p{1.5cm}<{\centering}}
		\toprule
		\multirow{2}*{Dataset} & \multirow{2}*{Method} & \multicolumn{4}{c}{Sampling Rates} \\
		& & $25\%$ & $50\%$ & $75\%$ & $100\%$ \\
		\midrule

		\multirow{5}*{CUB}     & Fine-tuning (baseline)            & $38.57\pm$0.13          & $58.97\pm$0.16          & $69.55\pm$0.18          & $74.35\pm$0.18    \\
		& GPT*~\citep{radford2019language}                & $36.43\pm$0.17          & $57.62\pm$0.14          & $67.82\pm$0.05          & $72.95\pm$0.29  \\
		& Center~\citep{DBLP:conf/eccv/WenZL016} 
		& $42.53\pm$0.41          & $62.15\pm$0.51          & $70.86\pm$0.39          & $75.61\pm$0.33   \\
		& BSS~\citep{Chen2019BSS}& $41.73\pm$0.14          & $59.15\pm$0.21          & $69.93\pm$0.19          & $74.16\pm$0.09	\\
		& \textbf{Bi-tuning}   & $\textbf{50.54}\pm$0.23 & $\textbf{66.88}\pm$0.13 & $\textbf{74.27}\pm$0.05 & $\textbf{77.14}\pm$0.23   \\
		\midrule
		\multirow{5}*{Cars}    & Fine-tuning (baseline)             & $62.40\pm$0.26          & $81.55\pm$0.36          & $88.07\pm$0.19          & $89.81\pm$0.48       \\
		& GPT*~\citep{radford2019language}                   & $65.83\pm$0.27          & $82.39\pm$0.17          & $88.62\pm$0.11          & $90.56\pm$0.18       \\
		& Center~\citep{DBLP:conf/eccv/WenZL016} 
		& $67.57\pm$0.12          & $82.78\pm$0.30          & $88.55\pm$0.24          & $89.95\pm$0.1  \\
		& BSS~\citep{Chen2019BSS}& $62.13\pm$0.22          & $81.72\pm$0.22          & $88.32\pm$0.17          & $90.41\pm$0.15     \\
		& \textbf{Bi-tuning}   & $\textbf{69.44}\pm$0.32 & $\textbf{84.41}\pm$0.07 & $\textbf{89.32}\pm$0.23 & $\textbf{90.88}\pm$0.13  \\
		\midrule
		\multirow{5}*{Aircraft}& Fine-tuning (baseline)             & $58.98\pm$0.54          & $77.39\pm$0.31          & $84.82\pm$0.24          & $87.35\pm$0.17 \\
		& GPT*~\citep{radford2019language}                   & $60.70\pm$0.08          & $78.93\pm$0.17          & $85.09\pm$0.10          & $87.56\pm$0.15        \\
		& Center~\citep{DBLP:conf/eccv/WenZL016} 
		& $62.23\pm$0.09          & $79.30\pm$0.14          & $85.20\pm$0.41          & $87.52\pm$0.20       \\
		& BSS~\citep{Chen2019BSS}& $60.13\pm$0.32          & $77.98\pm$0.29          & $84.85\pm$0.21          & $87.25\pm$0.07         \\
		& \textbf{Bi-tuning}   & $\textbf{63.16}\pm$0.26 & $\textbf{79.98}\pm$0.22 & $\textbf{86.23}\pm$0.29 & $\textbf{88.55}\pm$0.38  \\
		\bottomrule
	\end{tabular}

	\label{table:result-unsup}

\end{table*}

\begin{table*}[tbph]
	%\vspace{-10pt}
	\small
	\addtolength{\tabcolsep}{2pt}
	\centering
	\vskip 0.1in
	% \resizebox{\textwidth}{!}{%
	\caption{Top-1 Accuracy on \textbf{Car} dataset with different \textit{unsupervisedly pre-trained} representations.}
	
	\begin{tabular}{lcc}
		% {p{2.1cm}<{\centering}|c|p{1.5cm}<{\centering }p{1.5cm}<{\centering} p{1.5cm}<{\centering} p{1.5cm}<{\centering}}
		\toprule
		Pre-training Method    & Fine-tuning (100\% data) & \textbf{Bi-tuning} (100\%  data) \\
		\midrule
		Deep Cluster~\citep{caron2018deep}           & $83.90\pm$0.48 & \textbf{$\textbf{87.71}\pm$}0.34 \\ 
		InsDisc~\citep{wu2018unsupervised}           & $86.59\pm$0.22 & \textbf{$\textbf{89.54}\pm$}0.25 \\ 
		CMC~\citep{tian2019contrastive}              & $86.71\pm$0.62 & \textbf{$\textbf{88.35}\pm$}0.44 \\ 
		MoCov2~\citep{he2019momentum}                & $90.15\pm$0.48 & \textbf{$\textbf{90.79}\pm$}0.34 \\ 
		SimCLR($1\times$)~\citep{chen_simple_2020}   & $89.30\pm$0.18 & \textbf{$\textbf{90.84}\pm$}0.22 \\ 
		SimClR($2\times$)~\citep{chen_simple_2020}   & $91.22\pm$0.19 & \textbf{$\textbf{91.93}\pm$}0.19 \\ 
		
		\bottomrule
	\end{tabular}
	
	\label{table:result-pretraining}
	
\end{table*}
% \begin{table*}[h]
% 	%\vspace{-10pt}
% 	\addtolength{\tabcolsep}{2pt}
% 	\centering
% 	\vskip 0.1in
% 	% \resizebox{\textwidth}{!}{%
% 	\caption{Top-1 Accuracy on \% Car with differents pre-trained representations. Averaged over 5 trials.}
	
% 	\begin{tabular}{ccccc}
% 		% {p{2.1cm}<{\centering}|c|p{1.5cm}<{\centering }p{1.5cm}<{\centering} p{1.5cm}<{\centering} p{1.5cm}<{\centering}}
% 		\toprule
% 		Pre-training method & Fine-tuning (25\% data) & Bi-tuning (25\%  data)   & Fine-tuning (100\% data) & Bi-tuning (100\%  data) \\
% 		\midrule
% 		Deep Cluster~\citep{caron2018deep}           & $54.21\pm$0.42 & \textbf{$64.11\pm$}0.37 & $83.90\pm$0.48 & \textbf{$87.71\pm$0.34} \\ 
% 		InsDisc~\citep{wu2018unsupervised}           & $58.62\pm$0.19 & \textbf{$66.71\pm$}0.27 & $86.59\pm$0.22 & \textbf{$89.54\pm$0.25} \\ 
% 		CMC~\citep{tian2019contrastive}              & $56.71\pm$0.24 & \textbf{$64.23\pm$}0.13 & $86.71\pm$0.62 & \textbf{$88.35\pm$0.44} \\ 
% 		MoCov2~\citep{he2019momentum}                & $63.43\pm$0.42 & \textbf{$70.11\pm$}0.37 & $90.15\pm$0.48 & \textbf{$90.79\pm$0.34} \\ 
% 		SimCLR($1\times$)~\citep{chen_simple_2020}   & $62.55\pm$0.11 & \textbf{$69.63\pm$}0.14 & $89.30\pm$0.18 & \textbf{$90.84\pm$0.22} \\ 
% 		SimClR($2\times$)~\citep{chen_simple_2020}   & $66.42\pm$0.21 & \textbf{$71.23\pm$}0.17 & $91.22\pm$0.19 & \textbf{$91.93\pm$0.19} \\ 
		
% 		\bottomrule
% 	\end{tabular}
	
% 	\label{table:result-pretraining}
	
% \end{table*}

%\begin{figure*}[tbp]
\begin{figure*}[h]
	\centering
	\subfigure[Number of Sampling Keys ($K$)]{
		\includegraphics[width=0.48\textwidth]{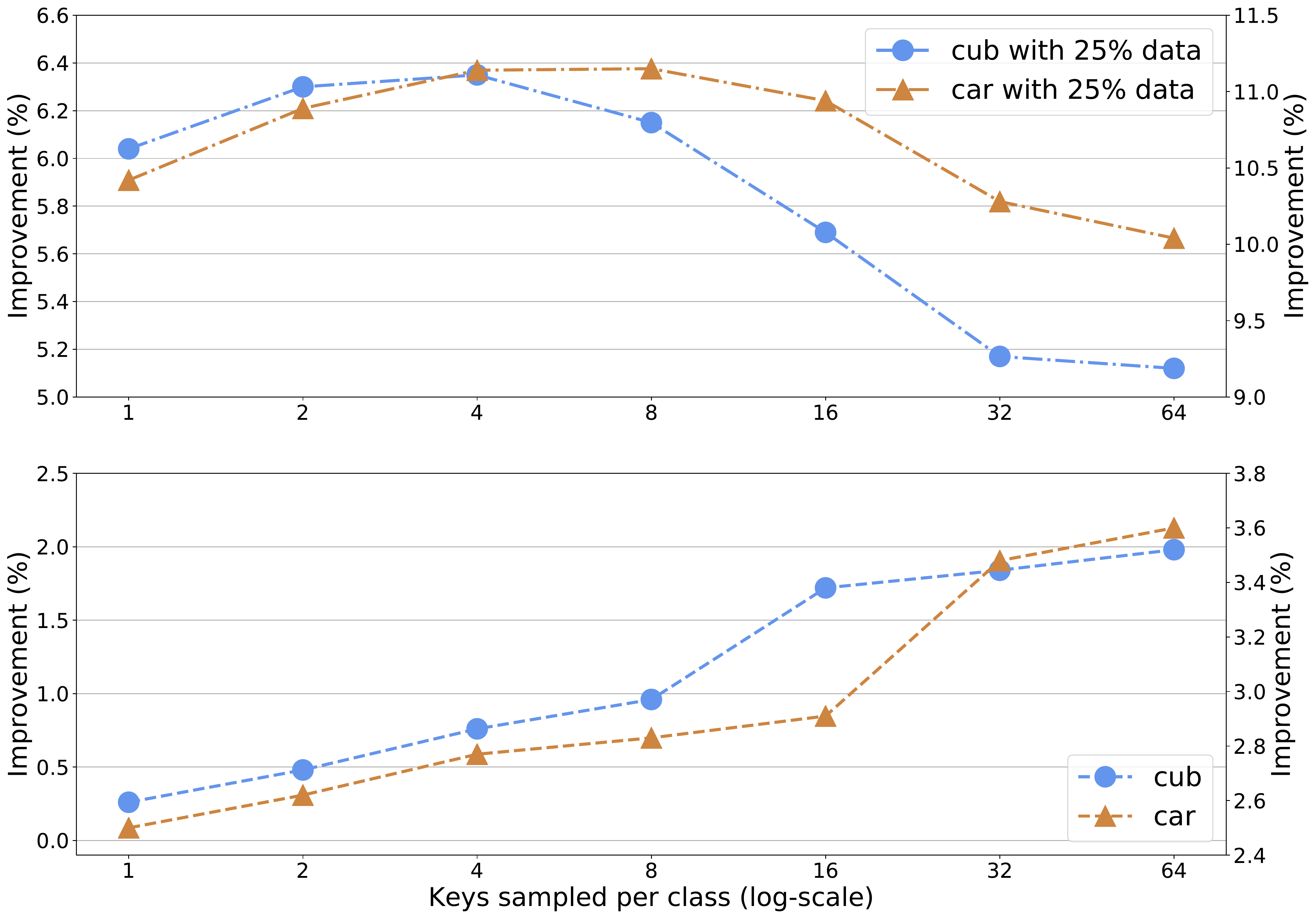}
		\label{fig:abk}
	}
	\subfigure[Dimension of Projector Head ($L$)]{
		\includegraphics[width=0.48\textwidth]{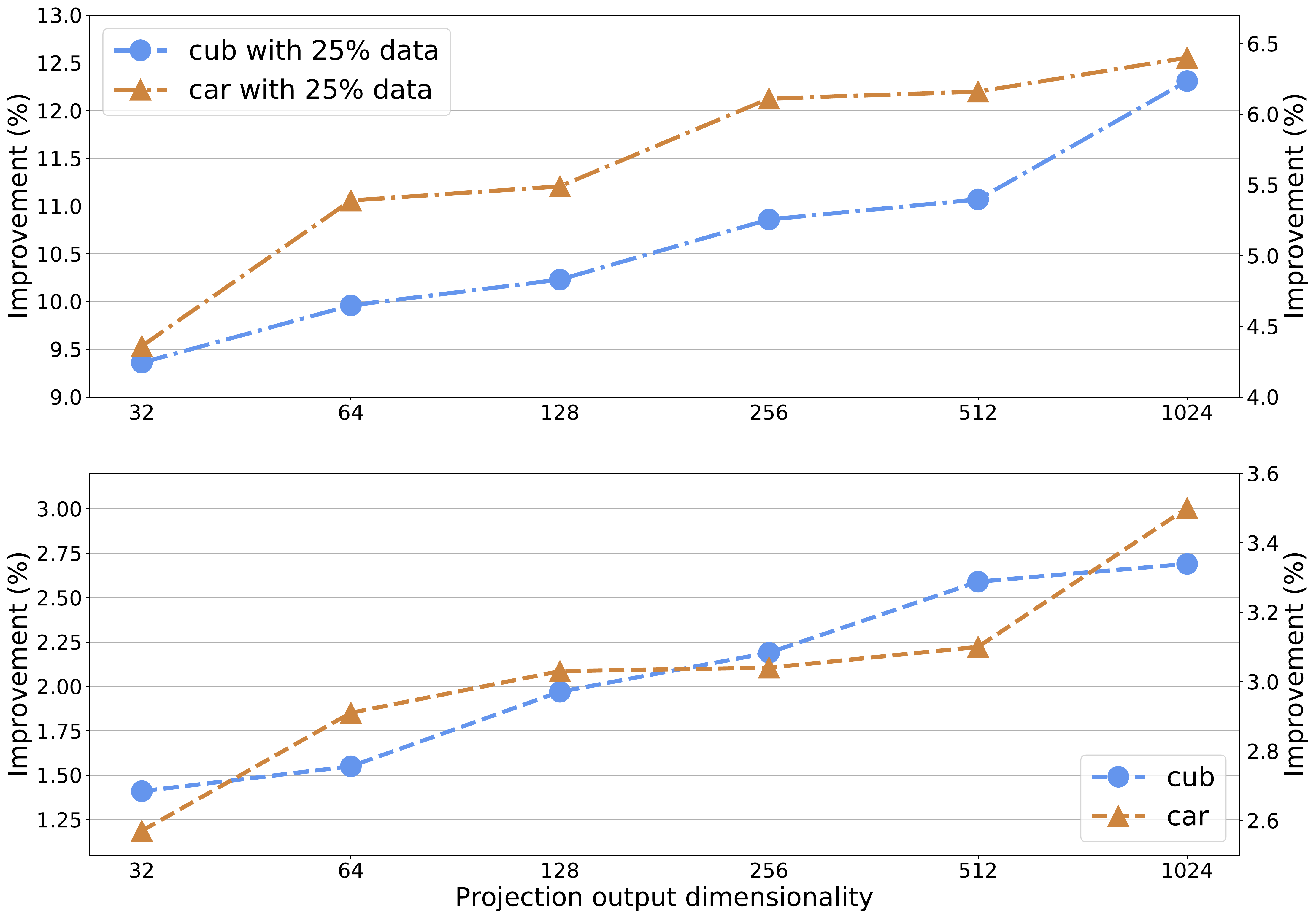}
		\label{fig:abdim}
	}
	\vspace{-10pt}
	\caption{Sensitivity analysis of hyper-parameters $K$ and $L$ for Bi-tuning.}
	\label{fig:ablation}
	\vspace{-5pt}
\end{figure*}

\begin{figure*}[h]
	\centering
	\subfigure[Original]{
		\includegraphics[width=0.18\textwidth]{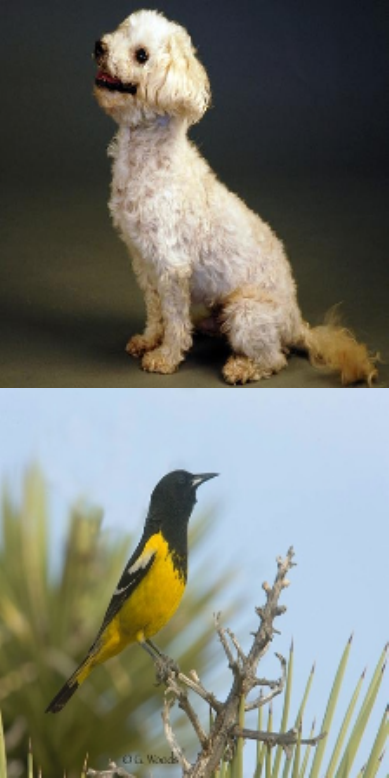}
		\label{fig:origin}
	}
	\subfigure[Random]{
		\includegraphics[width=0.18\textwidth]{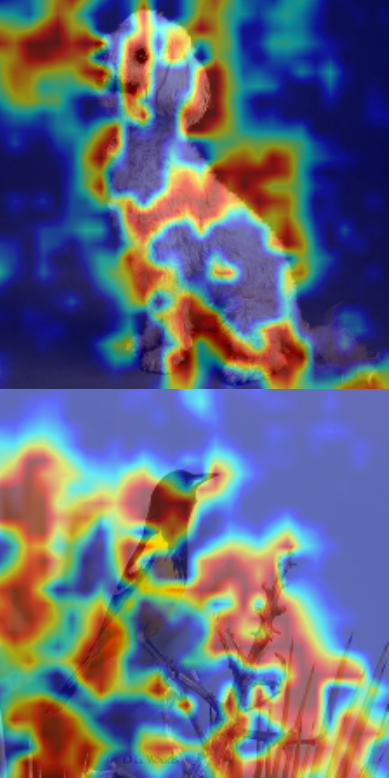}
		\label{fig:random}
	}
	\subfigure[Supervised]{
		\includegraphics[width=0.18\textwidth]{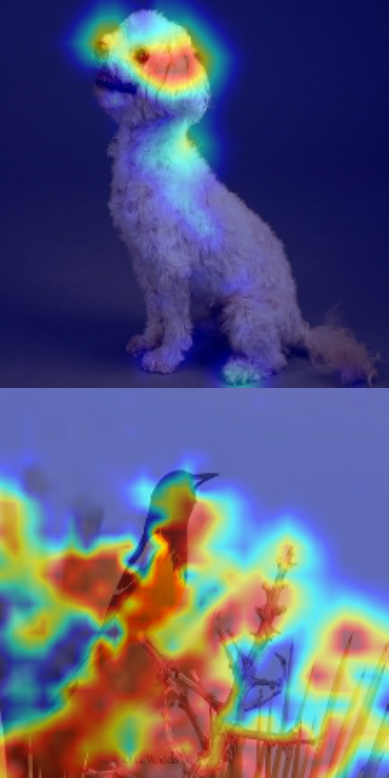}
		\label{fig:super}
	}
	\subfigure[Unsupervised]{
		\includegraphics[width=0.18\textwidth]{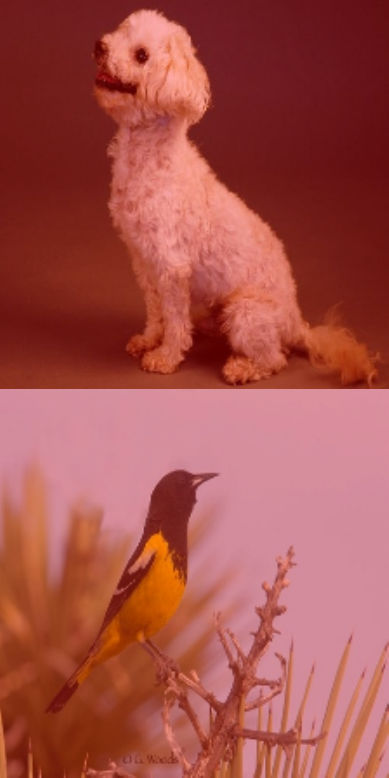}
		\label{fig:unsup}
	}
	\subfigure[Bi-tuning]{
		\includegraphics[width=0.18\textwidth]{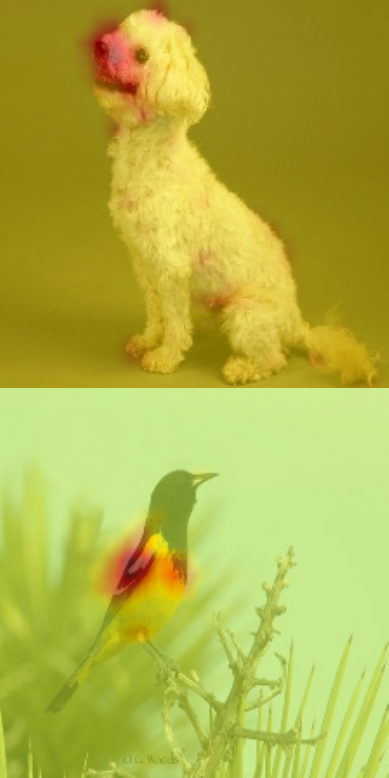}
		\label{fig:fine-tuning}
	}
	%\vspace{-5pt}
	\vspace{-10pt}
	\caption{Interpretable visualization of learned representations via various training methods. 
		%		We train a linear classifier on \textit{Dogs} \cite{pets-parkhi2012cats} for models that miss original pretext classifiers. 
		%		\ref{fig:l2sp} shows the $L_2$ distance between the original pre-trained parameters with the fine-tuning ones on both supervised and unsupervised pre-trained models, revealing a larger gap between the upstream unsupervised models with the downstream classification tasks.
	}
	\label{fig:motivation}
	\vspace{-5pt}
\end{figure*}

\subsection{Bi-tuning Unsupervised Pre-trained Representations}

\textbf{Bi-tuning representations of MoCo \citep{he2019momentum}.} In this round, we use ResNet-50 pre-trained unsupervisedly via {MoCo} on ImageNet as the backbone. Since suffering from the large discrepancy between unsupervised pre-trained representations and downstream classification tasks as demonstrated in Figure~\ref{fig:motivation}, previous fine-tuning competitors usually perform very poorly. Hence we only compare \textit{Bi-tuning} to the state-of-the-art method \textit{BSS}~\citep{Chen2019BSS} and vanilla fine-tuning as baselines. Besides, we add two intuitively related baselines: (1) \textbf{GPT*}, which follows a GPT ~\citep{radford2018improving,radford2019language} fine-tuning style but replaces its predictive loss with the contrastive loss; (2) \textbf{Center} loss, which introduces compactness of intra-class variations ~\citep{DBLP:conf/eccv/WenZL016} that is effective in recognition tasks. As reported in Figure~\ref{table:result-unsup}, trivially borrowing fine-tuning strategy in GPT~\citep{radford2018improving} or center loss brings tiny benefits, and is even harmful on some datasets, \textit{e.g.} \textbf{CUB}. Bi-tuning yields consistent gains on all fine-tuning tasks of unsupervised representations, indicating that Bi-tuning benefits substantially from exploring the intrinsic structure. 
%with a small optimal coefficient of tradeoff($0.01$ while $1$ in Bi-tuning).

\textbf{Bi-tuning other unsupervised pre-trained representations.} To justify Bi-tuning's general efficacy, we extend our method to unsupervised representations by other pre-training methods. Bi-tuning is applied to MoCo (version 2) \citep{he2019momentum}, SimCLR~\citep{chen_simple_2020}, InsDisc~\citep{wu2018unsupervised}, Deep Cluster~\citep{caron2018deep}, CMC~\citep{tian2019contrastive} on \textbf{Car} dataset with 100\% training data. Table~\ref{table:result-pretraining} is a strong signal that Bi-tuning is not bound to specific pre-training pretext tasks. 
%(ResNet50 1x, 2x models, corresponding to the "$\times 8$", "$\times 16$" cases in ~\cite{kolesnikov2019revisiting} because the standard-sized ResNet is referred to as "$\times 4$" in ~\cite{kolesnikov2019revisiting}).

\textbf{Analysis on components of contrastive learning.} 
%To implement CCE and CCL, a key sampler with high efficiency is necessary. 
Recent advances in contrastive learning, \emph{i.e.} momentum contrast~\citep{he2019momentum} and memory bank~\citep{wu2018unsupervised} can be plugged into Bi-tuning smoothly to achieve similar performance and the detailed discussions are deferred to \textit{Appendix}. 
% #key
Previous works ~\citep{he2019momentum,chen_simple_2020} reveal that a large amount of contrasts is crucial to contrastive learning. In Figure~\ref{fig:abk}, we report the sensitivity of the numbers of sampling keys in Bi-tuning (MoCo) under 25\% and 100\% sampling ratio configurations. 
%Note that CUB has various categories with a few images in each category. We let $K$ balancedly sampled from every category to simplify our analysis here.
 Figure~\ref{fig:abk} shows that though a larger key pool is beneficial, we cannot expand the key pool due to the limit of training data, which may lose sampling stochasticity during training. This result suggests that there is a trade-off between stochasticity and a large number of keys. \cite{chen_simple_2020} pointed out that the dimension of the projector also has a big impact. 
 The sensitivity of the dimension of projector head is also presented in Figure~\ref{fig:abdim}. 
 Note that the unsupervised pre-trained model (\textit{e.g.}, MoCo) may provide an off-the-shelf projector, fine-tuning or re-initializing it is almost the same ($90.88$ vs. $90.78$ on \textbf{Car} when $L$ is $128$).
% Mechanism

\textbf{Interpretable visualization of learned representations.}  As visualized by \cite{fong2017interpretable} shown in Figure~\ref{fig:motivation}. Note that \ref{fig:origin} is the original image, \ref{fig:random}, \ref{fig:super} and \ref{fig:unsup} are respectively obtained from a randomly initialized model, a supervised pre-trained model on ImageNet, and an unsupervised pre-trained model via MoCov1~\citep{he2019momentum}. The visualization shows that MoCo focuses only on local details.
%	 It is impressive that MoCo concentrates on all local details, indicating that MoCo may be an excellent initial model for localization tasks. 
In contrast, Bi-tuning in \ref{fig:fine-tuning} captures both local details and global category-structures.

% \subsection{Natural Language Processing}

% As a general fine-tuning framework, Bi-tuning's application boundary is also explored in NLP areas. We fine-tune BERT-base pre-trained model\cite{devlin2018bert} under the Bi-tuning framework to text classification and token classification tasks. Compare to vallina fine-tuning baseline, Bi-tuning still demonstrates benefits in NLP tasks.

% \begin{table*}[tbp]
% 	\addtolength{\tabcolsep}{2pt}
% 	\caption{Results of Bi-tuning in NLP.}
% 	\centering
% 	\begin{tabular}{ccccccc}
% 	  \toprule
% 		\multirow{2}*{Method} & \multicolumn{3}{c}{Text Classification} & \multicolumn{3}{c}{Token Classification} \\
% 		& SST-2 & TASK & TASK & TASK & ConN & GermNRE \\
% 	  \midrule
% 	  Fine-tuning & ??.?? & ??.??  & ??.??  & ??.??        & ??.?? &     \\
% 	  baseline-2  & ??.?? & ??.??  & ??.??  & ??.??        & ??.?? &     \\

% 	  Bi-tuning & ??.?? & ??.??  & ??.??  & ??.??        & ??.?? &     \\
% 	  \bottomrule
	  
%  	\end{tabular}
 
% 	\label{table:result-nlp}
% \end{table*}

\subsection{Collaborative Effect of Loss Functions}
\label{resandana}

%Results in Table~\ref{table:colla} are conducted to explore the collaborative effects among CE, CCE, and CCL, with choosing some different loss combinations. All experiments in Table~\ref{table:colla} are performed in CUB200 dataset~\citep{cub200-WelinderEtal2010} with ResNet-50 pre-trained via MoCo~\citep{he2019momentum}. To alleviate the gradient scale changing caused by removing one or two loss functions, we will rescale the final loss (e.g., if removing one loss, the left two will be multiplied by $1.5$). 
Using either contrastive cross-entropy (CCE) or categorical contrastive (CCL) with vanilla cross-entropy (CE) already achieves relatively good results, as shown in Table~\ref{table:colla}. These experiments prove that there is collaborative effect between CCE and CCL loss empirically. It is worth mentioning that CCE and CCL can work independently of CE, while we optimize these three losses simultaneously to yield the best result. As discussed in prior sections, we hypothesize that Bi-tuning helps fine-tuning models characterize the intrinsic structure of training data when using CCE and CCL simultaneously.

\begin{table*}[h]
	\addtolength{\tabcolsep}{2pt}
	\centering
	\caption{Collaborative effect in Bi-tuning on CUB-200-2011 using ResNet-50 pre-trained by MoCo.}
	\begin{tabular}{ccccccc}
		\toprule
		\multicolumn{3}{c}{Loss Function}  & \multicolumn{4}{c}{Sample Rate} \\
		CE & CCE & CCL & $25\%$ & $50\%$ & $75\%$ & $100\%$\\
		\midrule
		\ding{51} & \ding{55}& \ding{55} &$38.57\pm$0.13 &$58.97\pm$0.16 &$69.55\pm$0.18 & $74.35\pm$0.18\\
		\ding{51} & \ding{51}& \ding{55} &$45.42\pm$0.11 &$64.33\pm$0.28 &$71.56\pm$0.30 & $75.82\pm$0.21 \\
		\ding{51} & \ding{55}& \ding{51} &$41.09\pm$0.23 &$60.77\pm$0.31 &$70.30\pm$0.29 & $75.30\pm$0.20 \\
		\ding{55} & \ding{51}& \ding{51} &$47.70\pm$0.41 &$64.77\pm$0.15 &$71.69\pm$0.11 & $76.54\pm$0.24 \\
		\ding{51} & \ding{51}& \ding{51} &$\textbf{50.54}\pm$0.23 & $\textbf{66.88}\pm$0.13 & $\textbf{74.27}\pm$0.05 & $\textbf{77.12}\pm$0.23  \\
		\bottomrule
	\end{tabular}
	\label{table:colla}
\end{table*}

\section{Conclusion}
%Remarkable advances have been witnessed in unsupervised pre-trained models, but it remains unclear how to leverage the knowledge of unsupervised models to downstream classification tasks. 
In this paper, we propose a general Bi-tuning approach to fine-tuning both supervised and unsupervised representations. 
Bi-tuning generalizes the standard fine-tuning with an encoder for pre-trained representations, a classifier head and a projector head for exploring both the discriminative knowledge of labels and the intrinsic structure of data, which are trained end-to-end by two novel loss functions. 
Bi-tuning yields state-of-the-art results for fine-tuning tasks on both supervised and unsupervised pre-trained models by large margins. Code will be released upon publication at \url{http://github.com}.
%We justify through ablation studies the effectiveness of the proposed two-heads fine-tuning architecture with their novel loss functions.

%In this paper, we propose a novel {Co}ntrastive {Re}training (Bi-tuning) to fulfill this \textit{blank area }of deep learning community. Bi-tuning jointly model marginal and conditional distribution in a unified framework. Further, we propose a category-aware contrastive learning loss to tailor the commonly used unsupervised contrastive loss into a supervised retraining paradigm and a manifold-aware supervised learning loss to consider the manifold structure in the supervised retraining classifier. 

	\bibliography{ref/iclr2021}
	\bibliographystyle{iclr2021}

	\newpage
	\appendix
	
	\section{Visualization by t-SNE}
\begin{figure}[!htbp]
\centering
\subfigure[MoCo Representations]{
		\includegraphics[width=0.23\textwidth]{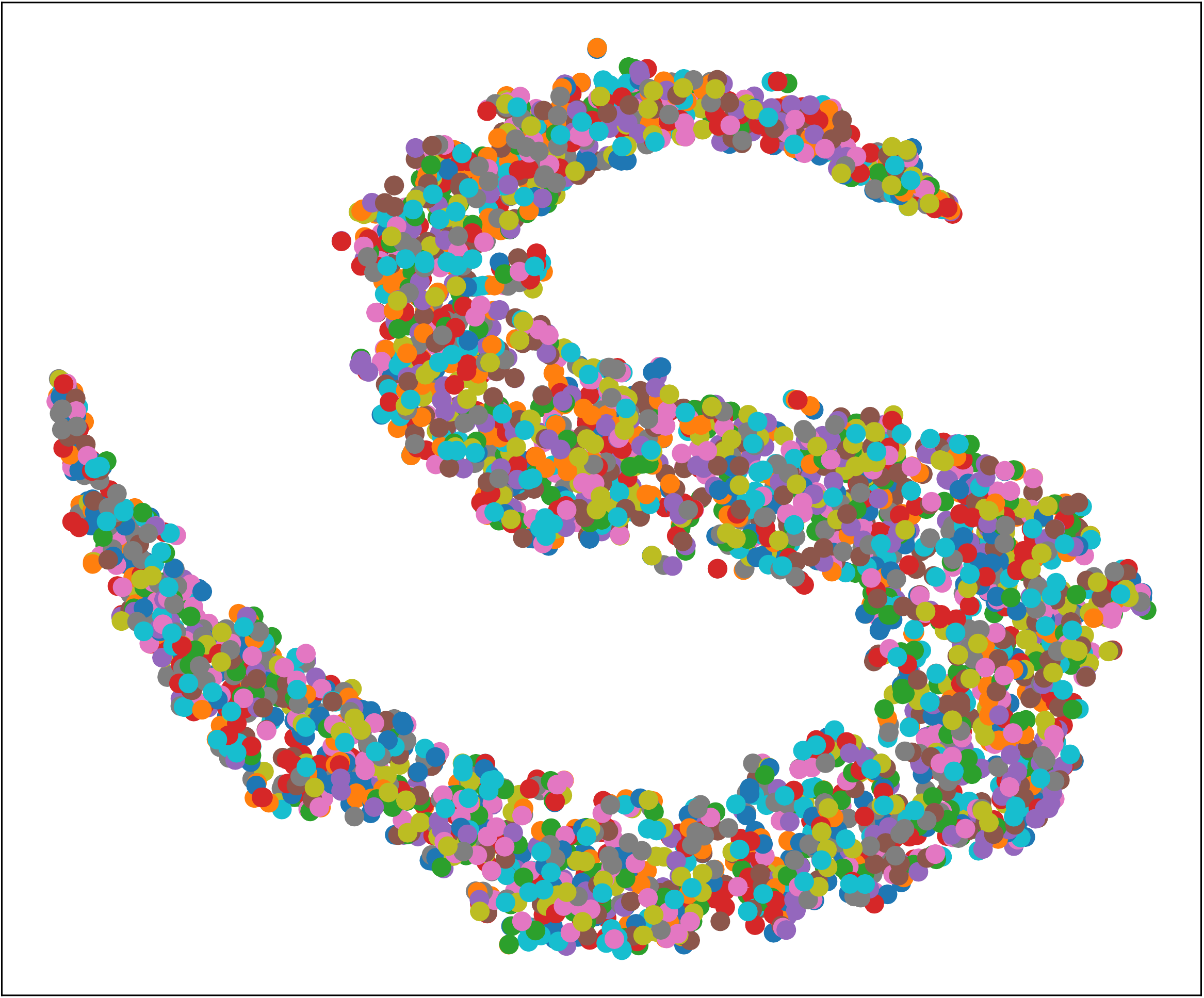}
		\label{fig:tsne-moco}
}
\subfigure[Fine-tuning with CE]{
		\includegraphics[width=0.23\textwidth]{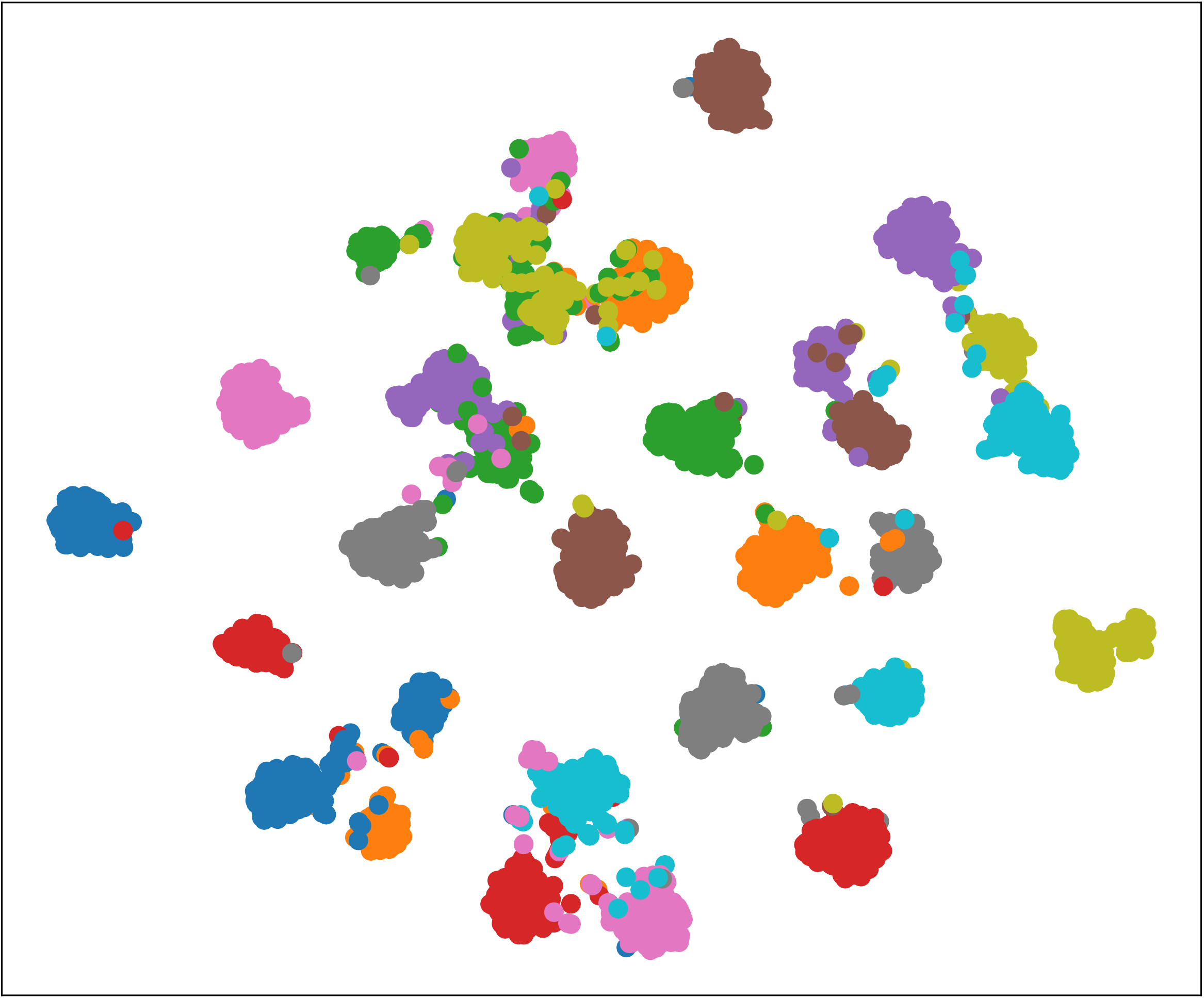}
		\label{fig:tsne-ce}
}
\subfigure[Fine-tuning with GPT*]{
		\includegraphics[width=0.23\textwidth]{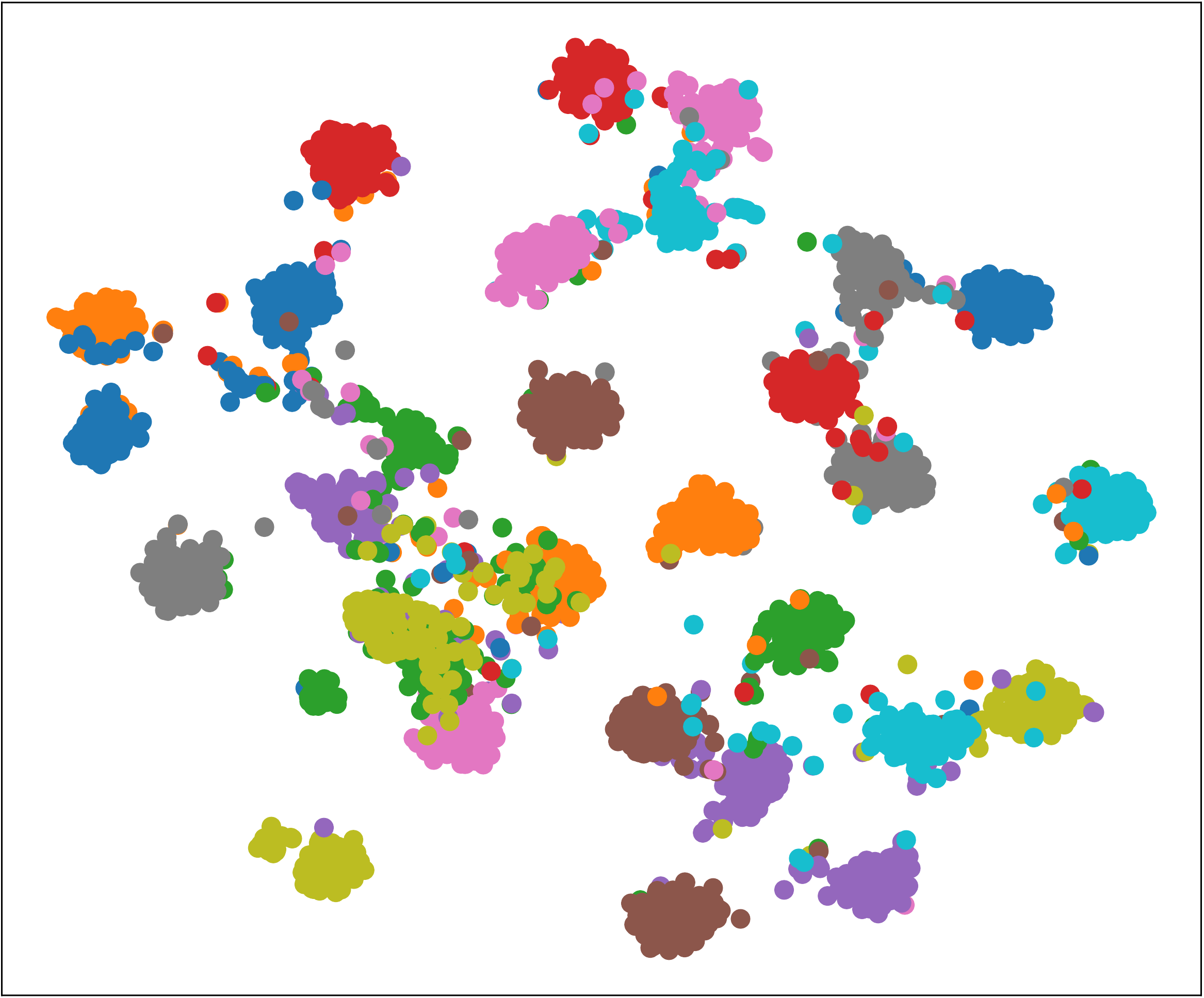}
		\label{fig:tsne-ce-bsl}
}
\subfigure[Bi-tuning]{
		\includegraphics[width=0.23\textwidth]{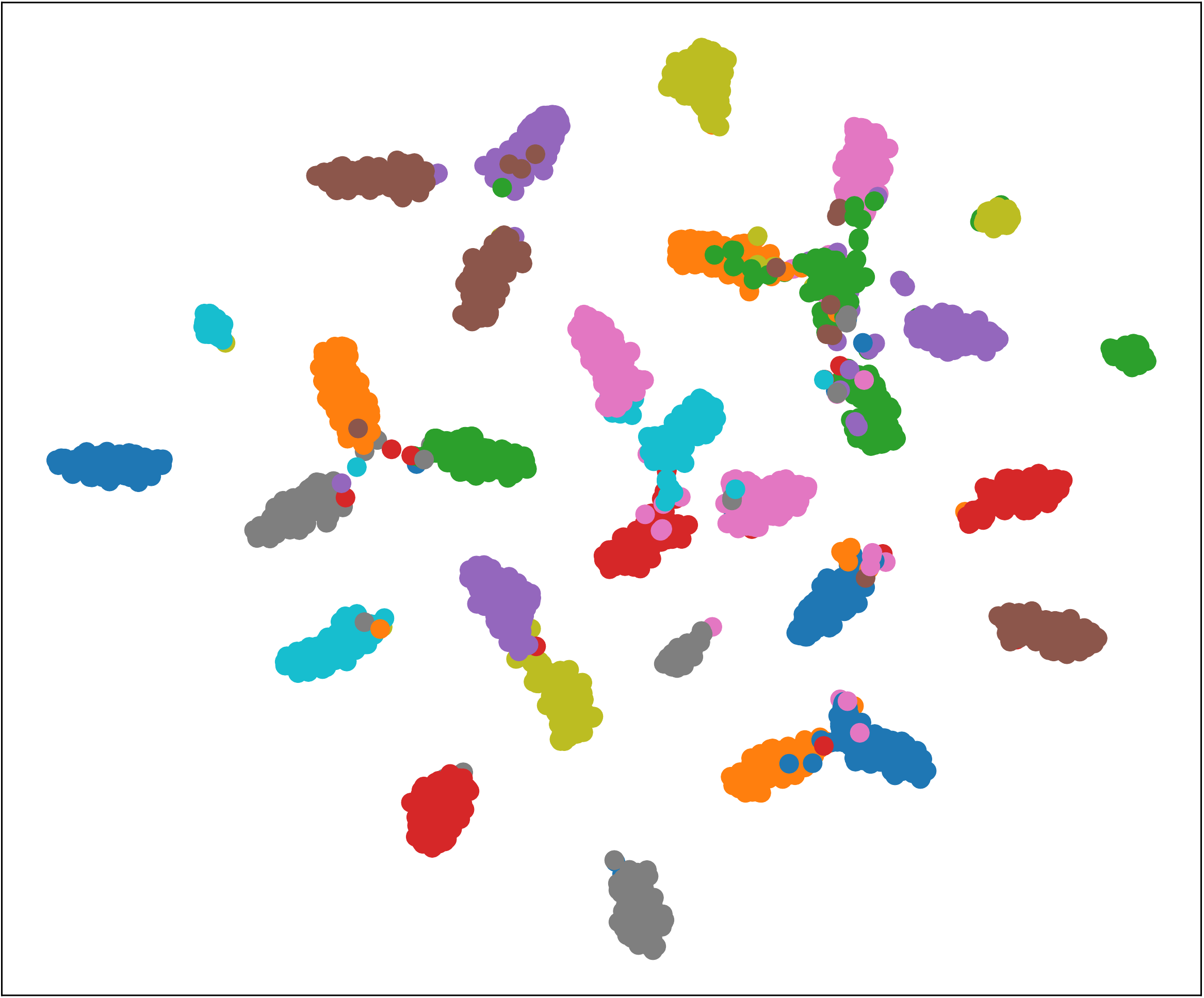}
		\label{fig:tsne-Bi-tuning}
}
\caption{ T-SNE~\cite{maaten2008visualizing} visualization of baselines on Pets~\cite{pets-parkhi2012cats}.}
\label{fig:TSNE}
\end{figure}

	 We train the t-SNE~\cite{maaten2008visualizing} visualization model on the MoCo representations fine-tuned on Pets dataset~\cite{pets-parkhi2012cats}. Visualization of the validation set is shown in Figure~\ref{fig:TSNE}. Note that representations in Figure~\ref{fig:tsne-moco} do not present good classification structures. Figure~\ref{fig:tsne-ce-bsl} suggests that forcefully combining the unsupervised contrastive learning loss as GPT~\cite{radford2019language} may cause conflict with CE and clutter the classification boundaries. Figure~\ref{fig:tsne-Bi-tuning} suggests Bi-tuning encourages the fine-tuning model to learn better intrinsic structure besides the label information. Therefore, Bi-tuning presents the best classification boundaries as well as intrinsic structures.
	 
\vspace{-10pt}

\section{Key Generating Mechanisms}

\subsection{Momentum Contrast}

Momentum Contrast (MoCo)~\cite{he2019momentum} is a general key generating mechanism for using contrastive loss. The main idea in MoCo is producing encoded keys on-the-fly via a momentum-updated encoder and maintaining a queue to support sampling operations. Thus, the memory cost in MoCo does not depend on the size of the training set (while a memory bank~\cite{wu2018unsupervised} will store the whole dataset). In all our experiments (Section 5), Bi-tuning chooses the unsupervised MoCo as our default setting.

Formally, denoting the momentum-updated encoder as $f_{\rm k}$ with parameters $\theta_{\rm k}$. Likewise, denoting the backbone encoder as $f_{\rm q}$ with parameters $\theta_{\rm q}$. $\theta_{\rm k}$ is updated by:
\begin{equation}\label{moco}
	\theta_{\rm k} \leftarrow m\theta_{\rm k} + (1-m)\theta_{\rm q}.
\end{equation}

Here we set the momentum coefficient $m=0.999$. To fit the Bi-tuning framework, we reorganize the queues in MoCo for items in each category separately. Moreover, two contrastive mechanisms in Bi-tuning are performed on different the instance level and category level respectively, and we maintain two groups of queues correspondingly.

\begin{table*}[h]
	\addtolength{\tabcolsep}{2pt}
	\centering
	\caption{Top-1 accuracy (\%) of Bi-tuning on CUB with memory bank as key generating mechanism (Backbone: ResNet-50 pretrained via MoCo).}
	\label{table:mecha}
	\begin{tabular}{lcccccc}
	  \toprule
	  \multirow{2}*{Key Generating Mechanism} & \multicolumn{4}{c}{Sample Rate} \\
	    & $25\%$ & $50\%$ & $75\%$ & $100\%$\\
	  \midrule
	  MoCo~\cite{he2019momentum}        & $49.25\pm$0.23 & $66.88\pm$0.13 & $74.27\pm$0.05 & $77.12\pm$0.23  \\
	  Memory bank~\cite{wu2018unsupervised} & $50.01\pm$0.55 & $66.69\pm$0.26 & $74.22\pm$0.31 & $77.62\pm$0.29  \\
	  \bottomrule
 	\end{tabular}
\end{table*}

\subsection{Memory Bank}

Bi-tuning is a general framework, which is not bound to any special key generating mechanism (MoCo). The memory bank proposed by \cite{wu2018unsupervised} generates encoded keys via momentum-updated snapshots of all items in the training set. Keys for each mini-batch are uniformly sampled from the memory bank. Compared to MoCo, maintaining a memory bank is more computation-efficient with more memory required. Similar to Eq. \eqref{moco}, snapshots here are updated by:
\begin{equation}\label{mb_z}
	\mathbf{z}_i^{\rm k} \leftarrow m\mathbf z_i^{\rm k} + (1-m)\mathbf z^{\rm q}_i,
\end{equation}
\begin{equation}\label{mb_h}
	\mathbf{h}_i^{\rm k} \leftarrow m\mathbf h_i^{\rm k} + (1-m)\mathbf h^{\rm q}_i.
\end{equation}

Notations follow Section 3. Here we set the momentum coefficient $m=0.5$ \cite{wu2018unsupervised}.  Other hyper-parameters are the same as Section 5. We evaluate Bi-tuning with a memory bank on CUB~\cite{cub200-WelinderEtal2010} with the same configurations in Section 4. The results in Table \ref{table:mecha} show that the performance is close in both methods. Key generating mechanisms in Bi-tuning only have limited effects on the final performance in the supervised paradigm. These suggest that the key generating mechanism in Bi-tuning can be implemented by some variants with similar performance. MoCo is recommended regarding its scalability and simplicity.

\end{document}